\pdfoutput=1

\documentclass[11pt]{article}

\usepackage{acl2022}

\usepackage{times}
\usepackage{latexsym}

\usepackage[T1]{fontenc}

\usepackage[utf8]{inputenc}

\usepackage{microtype}
\usepackage{graphicx}

\usepackage{multirow}
\usepackage{threeparttable}
\usepackage{comment}
\usepackage{graphicx}
\usepackage{makecell}
\usepackage{comment}
\usepackage{latexsym}
\usepackage{amsmath}
\usepackage{float}
\usepackage{rotating}
\usepackage{xcolor}
\usepackage{bbm}

\definecolor{c_05}{rgb}{0.9, 0.9, 1}
\definecolor{c_1}{rgb}{0.86, 0.86, 1}
\definecolor{c_15}{rgb}{0.82, 0.82, 1}
\definecolor{c_2}{rgb}{0.77, 0.77, 1}
\definecolor{c_23}{rgb}{0.75, 0.75, 1}
\definecolor{c_3}{rgb}{0.7, 0.7, 1}
\definecolor{c_4}{rgb}{0.66, 0.66, 1}
\definecolor{c_45}{rgb}{0.62, 0.62, 1}

\usepackage[mathscr]{euscript}
\usepackage[ruled,vlined, linesnumbered]{algorithm2e}
\SetKw{KwBy}{by}

\usepackage{amsmath,amssymb}
\usepackage{comment}

\usepackage{array}
\newcommand{\PreserveBackslash}[1]{\let\temp=\\#1\let\\=\temp}
\newcolumntype{C}[1]{>{\PreserveBackslash\centering}p{#1}}

\title{Domain Adaptation in Multilingual and Multi-Domain Monolingual Settings for Complex Word Identification}

\author{George-Eduard Zaharia, Răzvan-Alexandru Smădu \\ 
  \textbf{Dumitru-Clementin Cercel, Mihai Dascalu} \\
  University Politehnica of Bucharest, Faculty of Automatic Control and Computers\\
  \tt  \{george.zaharia0806, razvan.smadu\}@stud.acs.upb.ro\\
  \tt \{dumitru.cercel, mihai.dascalu\}@upb.ro\\
 }

\begin{document}
\maketitle
\begin{abstract}
Complex word identification (CWI) is a cornerstone process towards proper text simplification. CWI is highly dependent on context, whereas its difficulty is augmented by the scarcity of available datasets which vary greatly in terms of domains and languages. As such, it becomes increasingly more difficult to develop a robust model that generalizes across a wide array of input examples. In this paper, we propose a novel training technique for the CWI task based on domain adaptation to improve the target character and context representations. This technique addresses the problem of working with multiple domains, inasmuch as it creates a way of smoothing the differences between the explored datasets. Moreover, we also propose a similar auxiliary task, namely text simplification, that can be used to complement lexical complexity prediction. Our model obtains a boost of up to 2.42\% in terms of Pearson Correlation Coefficients in contrast to vanilla training techniques, when considering the CompLex from the Lexical Complexity Prediction 2021 dataset. At the same time, we obtain an increase of 3\% in Pearson scores, while considering a cross-lingual setup relying on the Complex Word Identification 2018 dataset. In addition, our model yields state-of-the-art results in terms of Mean Absolute Error.
\end{abstract}

\section{Introduction}

The overarching goal of the complex word identification (CWI) task is to find words that can be simplified in a given text ~\cite{paetzold-specia-2016-semeval}.
Evaluating word difficulty represents one step towards achieving simplified, which in return facilitates access to knowledge to a wider audience  texts~\cite{maddela2018wordcomplexity}. However, complex word identification is a highly contextualized task, far from being trivial. The datasets are scarce and, most of the time, the input entries are limited or cover different domains/areas of expertise. Therefore, developing a robust and reliable model that can be used to properly evaluate the complexity of tokens is a challenging task. Table~\ref{tab:examples} showcases examples of complex words annotations from the CompLex LCP ~\cite{shardlow2020complex, shardlow2021predicting} and English CWI ~\cite{yimam2018report} datasets employed in this work.

\begin{table}[t]
\centering
\resizebox{0.48\textwidth}{!}{\renewcommand{\arraystretch}{1.5}
\begin{tabular}{|c|p{1.7cm}|m{8cm}|}
\hline
                  & \textbf{Domain} & \textbf{Text}                                                                                                            \\ \hline
\hline
\multirow{3}{*}[-1em]{\begin{sideways}CompLex LCP Dataset\end{sideways}} & Bible           & But let each man test his \colorbox{c_05}{own work}, and then he will take pride in himself and not in his neighbor.                       \\ \cline{2-3} 
 &
  Biomedical &
  A genome \colorbox{c_2}{database search} revealed orthologs of ADAM11, ADAM22 and ADAM23 genes to exist in vertebrates such as mammals, fish, and amphibians, but not in invertebrates. \\ \cline{2-3} 
                  & Europarl        & They also allow for \colorbox{c_23}{easy compensation} for the thousands of accidents involving vehicles from more than one Member State. \\ \hline
\hline
\multirow{3}{*}[-1em]{\begin{sideways}English CWI Dataset \end{sideways}} &
  Wikipedia &
  \colorbox{c_05}{Normally}, the land will be \colorbox{c_05}{passed} down to \colorbox{c_15}{future generations} in a way that \colorbox{c_3}{recognizes} the community's \colorbox{c_2}{traditional} connection to that \colorbox{c_05}{country}. \\ \cline{2-3} 
 &
  WikiNews &
  The JAS 39C \colorbox{c_1}{Gripen} \colorbox{c_1}{crashed} onto a \colorbox{c_15}{runway} at around 9:30 am local time (02:30 UTC) and \colorbox{c_4}{exploded}, closing the airport to \colorbox{c_45}{commercial} flights. \\ \cline{2-3} 
                  & News            & The car has been \colorbox{c_05}{removed} from the scene for \colorbox{c_4}{forensic technical examination}.                                              \\ \hline
\end{tabular}}
\caption{\label{tab:examples} Examples of complex words annotated for each of the domains from CompLex LCP and CWI datasets. The shades indicate the complexity; the darker the shade, the more complex the sequence of words. Best viewed in color.}
\end{table}

Nevertheless, certain training techniques and auxiliary tasks help the model improve its generalization abilities, forcing it to focus only on the most relevant, general features~\cite{schrom2020improved}. Techniques like domain adaptation~\cite{ganin2016domainadversarial} can be used for various tasks, with the purpose of selecting relevant features for follow-up processes. At the same time, the cross-domain scenario can be transposed to a cross-lingual setup, where the input entries are part of multiple available languages. Performance can be improved by also employing the power of domain adaptation, where the domain is the language; as such, the task of identifying complex tokens can be approached even for low resource languages.

We propose several solutions to improve the performance of a model for CWI in a cross-domain or a cross-lingual setting, by adding auxiliary components (i.e., Transformer~\cite{vaswani2017attention} decoders, Variational Auto Encoders - VAEs~\cite{kingma2014autoencoding}), as well as a domain adaptation training technique~\cite{farahani2020brief}. Moreover, we use the domain adaptation intuition and we apply it in a multi-task adversarial training scenario, where the main task is trained alongside an auxiliary one, and a task discriminator has the purpose of generalizing task-specific features.

We summarize our main contributions as follows:

\begin{itemize}
  \item Applying the concept of domain adaptation in a monolingual, cross-domain scenario for complex word identification;
  \item Introducing the domain adaptation technique in a cross-lingual setup, where the discriminator has the purpose to support the model extract only the most relevant features across all languages;
  \item Proposing additional components (i.e., Transformer decoders and Variational Auto Encoders) trained alongside the main CWI task to provide more meaningful representations of the inputs and to ensure robustness, while generating new representations or by tuning the existing ones;
  \item Experimenting with an additional text simplification task alongside domain/language adaptation, with the purpose of extracting cross-task features and improving performance.
\end{itemize}

\section{Related Work}

\textbf{Domain Adaptation}. Several works employed domain adaptation to improve performance. For example, \newcite{du-etal-2020-adversarial} approached the sentiment analysis task by using a BERT-based~\cite{devlin2019bert} feature extractor alongside domain adaptation. Furthermore, ~\newcite{mchardy-etal-2019-adversarial} used domain adaptation for satire detection, with the publication source representing the domain. At the same time, \newcite{dayanik-pado-2020-masking} used a technique similar to domain adaptation, this time for political claims detection. The previous approaches consisted of actor masking, as well as adversarial debiasing and sample weighting. Other studies considering domain adaptation included suggestion mining~\cite{klimaszewski-andruszkiewicz-2019-wut}, mixup synthesis training~\cite{TangYuhua}, and effective regularization~\cite{vernikos2020domain}.

\textbf{Cross-Lingual Domain Adaptation}. \newcite{chen2018adversarial} proposed ADAN, an architecture based on a feed-forward neural network with three main components, namely: a feature extractor, a sentiment classifier, and a language discriminator. The latter had the purpose of supporting the adversarial training setup, thus covering the scenario where the model was unable to detect whether the input language was from the source dataset or the target one. A similar cross-lingual approach was adopted by \newcite{zhang-etal-2020-margin}, who developed a system to classify entries from the target language, while only labels from the source language were provided.

\newcite{keung-etal-2019-adversarial} employed the usage of multilingual BERT~\cite{pires2019multilingual} and argued that a language-adversarial task can improve the performance of zero-resource cross-lingual transfers. Moreover, training under an adversarial technique helps the Transformer model align the representations of the English inputs.

Under a Named Entity Recognition training scenario, \newcite{kim-etal-2017-cross} used features on two levels (i.e., word and characters), together with Recurrent Neural Networks and a language discriminator used for the domain-adversarial setup. Similarly, ~\newcite{huang-etal-2019-cross} used target language discriminators during the process of training models for low-resource name tagging.

\textbf{Word Complexity Prediction}. \newcite{gooding-kochmar-2019-complex} based their implementation for CWI as a sequence labeling task on Long Short-Term Memory (LSTM)~\cite{HochreiterLSTM} networks, inasmuch as the context helps towards proper identification of complex tokens. The authors used 300-dimensional pre-trained word embeddings as inputs for the LSTMs. Also adopting a sequence labeling approach, ~\newcite{finnimore-etal-2019-strong} considered handcrafted features, including punctuation or syllables, that can properly identify complex structures.

The same sequence labeling approach can be applied under a plurality voting technique~\cite{PolikarR}, or even using an Oracle~\cite{KunchevaL}. The Oracle functions best when applied to multiple solutions, by jointly using them to obtain a final prediction. At the same time, \newcite{zaharia2020crosslingual} explored the power of Transformer-based models~\cite{vaswani2017attention} in cross-lingual environments by using different training scenarios, depending on the scarcity of the resources: zero-shot, one-shot, as well as few-shot learning. Moreover, CWI can be also approached as a probabilistic task. For example, \newcite{de-hertog-tack-2018-deep} introduced a series of architectures that combine deep learning features, as well as handcrafted features to address CWI as a regression problem. 

\section{Method}

\subsection{Datasets}

We experimented with two datasets, one monolingual - CompLex LCP 2021~\cite{shardlow2020complex, shardlow2021predicting} - and one cross-lingual - the CWI Shared Dataset ~\cite{yimam2018report}. The entries of \textit{CompLex} consist of a sentence in English and a target token, alongside the complexity of the token, given its context. The complexities are continuous values between 0 and 1, annotated by various individuals on an initial 5-point Likert scale; the annotations were then normalized.

The \textit{CompLex} dataset contains two types of entries, each with its corresponding subset of entries: a) single, where the target token is represented by a single word, and b) multiple, where the target token is represented by a group of words. While the single-word dataset contains 7,662 training entries, 421 trial entries, and 917 test entries, the multi-word dataset has lower counts, with 1,517 training entries, 99 trial entries, and 184 for testing. At the same time, the entries correspond to three different domains (i.e., biblical, biomedical, and political), therefore displaying different characteristics and challenging the models towards generalization. 

The \textit{CWI dataset} was introduced in the CWI Shared Task 2018~\cite{yimam2018report}. It is a multilingual dataset, containing entries in English, German, Spanish, and French. Moreover, the English entries are split into three categories, depending on their proficiency levels: professional (News), non-professional (WikiNews), and Wikipedia articles. Most entries are for the English language (27,299 training and 3,328 validation), while the fewest training entries are for German (6,151 training and 795 validation). The French language does not contain training or validation entries.

\subsection{The Domain Adaption Model}
\begin{figure*}[htp]
\centering
\includegraphics[width=0.8\linewidth]{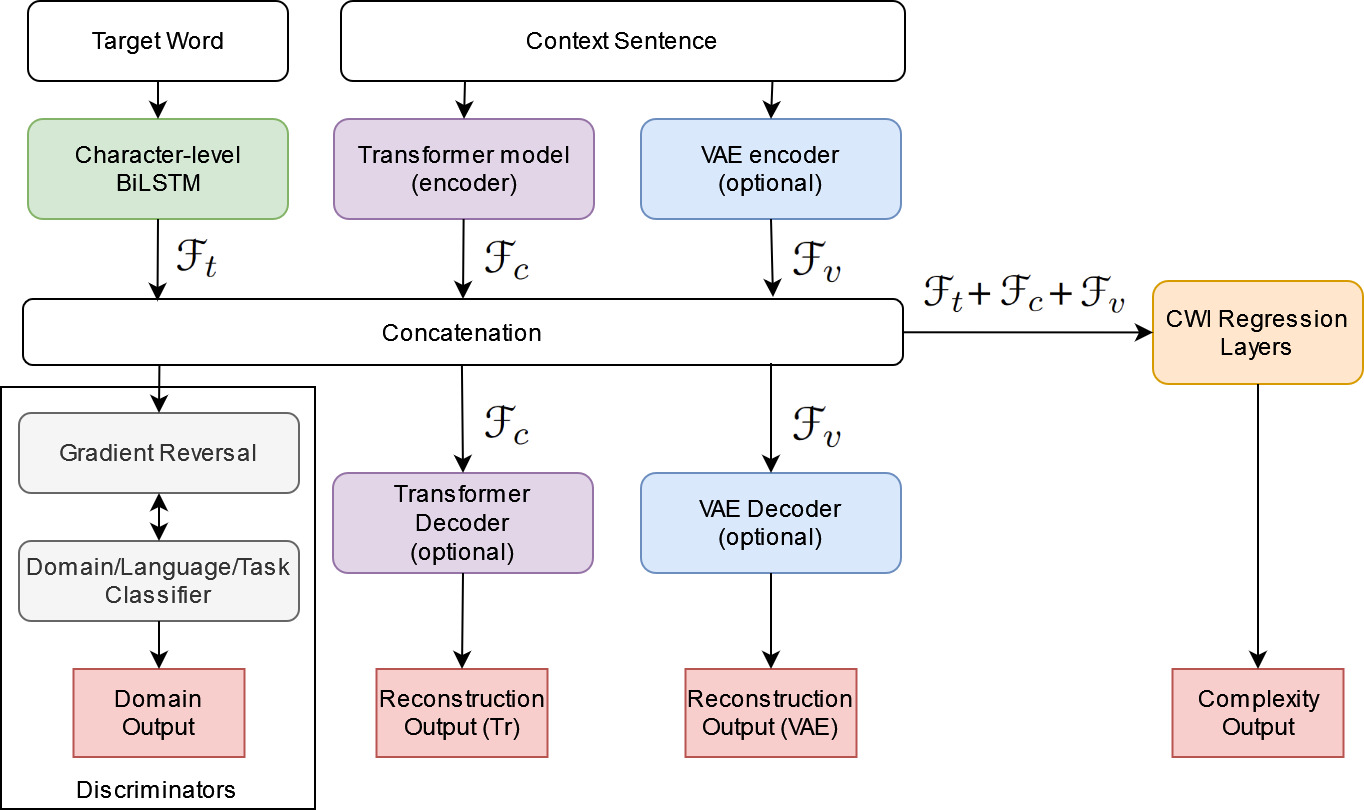}
  \caption{The overarching architecture for the domain adaptation model.}
  \label{fig:finalmodel}
\end{figure*}

The overarching architecture of our method is introduced in Figure~\ref{fig:finalmodel}. All underlying components are presented in detail in the following subsections. Our model combines character-level BiLSTM features (i.e., $ \mathscr{F}_t $) with Transformer-based features for the context sentence (i.e., $\mathscr{F}_c$). The concatenated features ($\mathscr{F}_c$+$ \mathscr{F}_t $) are then passed through three linear layers, with a dropout separating the first and second. The output is a value representing the complexity of the target word.

Three configurations were experimented. Within \textbf{Basic Domain Adaptation}, the previous features are passed through an additional component, the domain discriminator, composed of a linear layer followed by a softmax activation function. A \textit{gradient reversal} layer~\cite{ganin2015unsupervised} is added between the feature concatenation and the discriminator to reverse the gradients through the backpropagation phase and support extracting general features. The loss function is determined by Equation \ref{eq:daloss} as:
\begin{equation}
\label{eq:daloss}
  \begin{aligned}
      L &= L_r - \beta \lambda L_d 
  \end{aligned}
\end{equation}
where \textit{L\textsubscript{r}} is the regression loss, \textit{L\textsubscript{d}} is the general domain loss, $\beta$ is a hyperparameter used for controlling the importance of L\textsubscript{d}, and $ \lambda $ is another hyperparameter that varies as the training process progresses.

The following setups also include the Basic Domain Adaptation training setting.

\textbf{VAE and Domain Adaptation} considers the previous configuration, plus the VAE encoder, that yields the $\mathscr{F}_v$ features, and the VAE decoder, which aims to reconstruct the input. The concatenation layer now contains the BiLSTM and Transformer features, along with the VAE encoder features ($\mathscr{F}_v$), namely $\mathscr{F}_t$+$\mathscr{F}_c $+$\mathscr{F}_v$. The loss function is depicted by Equation~\ref{eq:vaedaloss} as:
\begin{equation}
\label{eq:vaedaloss}
  \begin{aligned}
      L &= L_r - \beta \lambda L_d + \alpha L_v
  \end{aligned}
\end{equation}
where, additionally, \textit{L\textsubscript{v}} represents the VAE loss described in Equation \ref{eq:vaeloss}.

\textbf{Transformer Decoder and Domain Adaptation} adds a Transformer Decoder with the purpose of reconstructing the original input, for a more robust context feature extraction. The loss is denoted by Equation \ref{eq:decoderdaloss} as:
\begin{equation}
\label{eq:decoderdaloss}
  \begin{aligned}
      L &= L_r - \beta \lambda L_d + \alpha L_{dec}
  \end{aligned}
\end{equation}
where \textit{L\textsubscript{dec}} represents the decoder loss described in Equation \ref{eq:decoderloss}.

\subsubsection{Character-level BiLSTM for Target Word Representation}
The purpose of this component is to determine the complexity of the target token, given only its constituent characters. A character-level Bidirectional Long Short-Term Memory (BiLSTM) network receives as input an array of characters corresponding to the target word (or group of words), and yields a representation that is afterwards concatenated to the previously mentioned Transformer-based representations. Each character \textit{c} is mapped to a certain value obtained from the character vocabulary \textit{V}, containing all the characters present in the input dataset.

The character sequence is represented as \textit{C\textsuperscript{i}} = \textit{[c\textsubscript{1}, c\textsubscript{2}, …, c\textsubscript{n}]}, where \textit{n} is the maximum length of a target token. \textit{C\textsuperscript{i}} is then passed through a character embedding layer, thus yielding the output \textit{Emb\textsubscript{target}}. \textit{Emb\textsubscript{target}} is then fed to the BiLSTM, followed by a dropout layer, thus obtaining the final target word representation, $ \mathscr{F}_t $.

\subsubsection{Transformer-based Context Representation}
We rely on a Transformer-based model as the main feature extractor for the context of the target word (i.e., the full sentence), considering their superior performance on most natural language processing tasks. The selected model for the first dataset is RoBERTa~\cite{liu2019roberta}, as it yields better results when compared to its counterpart, BERT. RoBERTa is trained with higher learning rates and larger mini-batches, and it modifies the key hyperparameters of BERT. We employed the usage of XLM-RoBERTa~\cite{conneau2020unsupervised}, the multilingual counterpart of RoBERTa, now trained on a very large corpus of multilingual texts, for the second cross-lingual task. The features used for our task are represented by the pooled output of the Transformer model. The feature vector $\mathscr{F}_c$ of 768 elements captures information about the context of the target word.

\subsubsection{Variational AutoEncoders}
We aim to further improve performance by adding extra features via Variational AutoEncoders (VAEs)~\cite{kingma2014autoencoding} to the context representation for a target word. More specifically for the CWI task, we use the latent vector \textit{z}, alongside the Transformer and the Char BiLSTM features. Moreover, we also need to ensure that the Encoder representation is accurate; therefore, we consider the VAE encoding and decoding as an additional task having the purpose of minimizing the reconstruction loss.

The VAE consists of two parts, namely the encoder and the decoder. The encoder \textit{g(x)} produces the approximation \textit{q(z|x)} of the posterior distribution \textit{p(z|x)}, thus mapping the input \textit{x} to the latent space \textit{z}. The process is presented in Equation~\ref{eq:vaeencoder}. We use as features the representation \textit{z}, denoted as $\mathscr{F}_v$.
\begin{equation}
\label{eq:vaeencoder}
  \begin{aligned}
      p(z|x) \approx q(z|x) = \mathcal{N}(\mu(x), \sigma(x)) 
  \end{aligned}
\end{equation}

The decoder \textit{f(z)} maps the latent space to the input space (i.e., \textit{p(z)} to \textit{p(x)}), by using Equation ~\ref{eq:vaedecoder}.
\begin{equation}
\label{eq:vaedecoder}
  \begin{aligned}
      p(x) &= \int p(x|z)p(z)dz \\
           &=\int \mathcal{N}(f(z),I)p(z)dz 
  \end{aligned}
\end{equation}

Equation \ref{eq:vaeloss} introduces the loss function, where \textit{D\textsubscript{KL}} represents the Kullback Leibler divergence. Furthermore, $ \mathbb{E} $\textsubscript{q} represents the expectation with relation to the distribution \textit{q}.
\begin{equation}
\label{eq:vaeloss}
  \begin{aligned}
      L(f,g) & = \sum_{i} \{-D_{KL}[q(z|x_i)||p(z)] \\
             & + \mathbb{E}_{q(z|{x_i})} [\ln p(x_i|z)]\}
  \end{aligned}
\end{equation}

\subsubsection{Discriminators}
The features extracted by our architecture can vary greatly as the input entries can originate from different domains or languages. Consequently, we introduced a generalization technique to extract only cross-domain features that do not present a bias towards a certain domain. We thus employ an adversarial training technique based on domain adaptation, forcing the model to only extract relevant cross-domain features.

A discriminator acts as a classifier, containing three linear layers with corresponding activation functions. The discriminator classifies the input sentence into one of the available domains. Unlike traditional classification approaches, our purpose is not to minimize the loss, but to maximize it. We want our model to become incapable of distinguishing between different categories of input entries, therefore extracting the most relevant, cross-domain features.

Our architecture is encouraged to generalize in terms of extracted features by the \textit{gradient reversal} layer that reverses the gradients during the backpropagation phase; as such, the parameters are updated towards the direction that maximizes the loss instead of minimizing it.

Three scenarios were considered, each one targeting a different approach towards domain adaptation.

\textbf{Domain Discriminator}. The first scenario is applied on the first dataset, CompLex, with entries only in English, but covering multiple domains. The discriminator has the purpose of identifying the domain of the entry, namely biblical, biomedical or political. The intuition is that, by grasping only cross-domain features, the performance of the model increases on all three domains, instead of performing well only on one, while poorer on the others.

\textbf{Language Discriminator}. The intuition is similar to the previous scenario, except that we experimented with the second multilingual dataset. Therefore, our interest was that our model extracts cross-lingual features, such that the performance is equal on all the target languages.

\textbf{Task Discriminator}. In this scenario, we trained a similar, auxiliary task, represented by text simplification. A task discriminator is implemented to detect the origin of the input entry: either the main task or the auxiliary task (i.e., simplified version). The dataset used for text simplification is represented by BenchLS~\cite{paetzold_2016_2552393}\footnote{\url{http://ghpaetzold.github.io/data/BenchLS.zip}}. The employed simplification process  consists of masking the word considered to be complex and then using a Transformer for Masked Language Modeling to predict the best candidate. The corresponding flow is described in Algorithm~\ref{mlmalgorithm}, while the loss function is presented in Equation~\ref{eq:multitaskdaloss}:
\begin{equation}
\label{eq:multitaskdaloss}
  \begin{aligned}
      L &= L_r - \beta \lambda L_{task\_id} + L_{ML}
  \end{aligned}
\end{equation}
where \textit{L\textsubscript{ML}} is the Sparse Categorical Cross Entropy loss.

\begin{algorithm}[!t]
\SetAlgoLined
 \textbf{Inputs:} Preprocessed dataset, split into batches (\textit{x\textsubscript{i}}, \textit{y\textsubscript{i}}), \textit{i=1,n} (where \textit{n} is the number of batches, \textit{x\textsubscript{i}} are the input features for the target word and the context, and \textit{y\textsubscript{i}} is the complexity)\;
 \textbf{Outputs:} Updated parameters $ \theta_p $\;

\textbf{Initialization:} Initialize $\theta_p$ with random weights\;
\For{every batch}{
    Select entries E1 from Task 1\;
    Select entries E2 from Task 2\;
    out1 = Apply initial architecture on E1\;
    out2 = Apply Masked Language Modeling Transformer on E2\;
    F = Combine the features from applying architecture on E1 and E2\;
    out\_task = Pass F through task discriminator\;
    
    loss1 = L\textsubscript{r}(out1, ref1)\;
    loss2 = L\textsubscript{ML}(out, ref2)\;
    task\_loss = L\textsubscript{task\_id}(out\_task, ref\_task)\;
    loss = loss1+loss2-$\beta \lambda$task\_loss\;
    Backpropagate loss\;
    Update $ \theta_p $\;
}

 \caption{The Multi-Task Adversarial algorithm (Task 1 - lexical complexity prediction; Task 2 - text simplification).}\label{mlmalgorithm}
\end{algorithm}

All previous discriminators use the same loss, namely Categorical Cross Entropy~\cite{zhang2018generalized}.

The overall loss consists of the difference between the task loss and the domain/language loss. Moreover, the importance of the latter can be controlled by multiplication with a $\lambda$ hyperparameter, that changes over time, and a fixed $\beta$ hyperparameter. The network parameters, $\theta_p$ are updated according to Equation~\ref{eq:paramsupdate}, where $\eta$ is the learning rate, $L_d$ is the domain loss, $L_r$ is the task loss and $ \beta $ is the weight for the domain loss. A similar equation for language loss ($L_l$) is in place for the second dataset, where instead of the domain loss $L_d$ we used the language identification loss $L_l$, having the same formula.

\begin{equation}
\label{eq:paramsupdate}
  \begin{aligned}
     \theta_p = \theta_p - \eta (\frac{\partial L_r}{\partial \theta_p} - \beta\lambda\frac{\partial L_d}{\partial \theta_p})
  \end{aligned}
\end{equation}

\subsubsection{Transformer Decoder}
Our model also considers a decoder to reconstruct the original input, starting from the Transformer representation. The intuition behind introducing this decoder is to increase the robustness of the context feature extraction.

The decoder receives as input the outputs of the hidden Transformer layer alongside an embedding of the original input, which are passed through a Gated Recurrent Unit (GRU)~\cite{chung2014empirical} layer for obtaining the final representation of the initial input. Additionally, two linear layers separated by a dropout are introduced before obtaining the final representation, \textit{y} = $\mathscr{F}_d$. The loss is computed by using the Negative Log Likelihood loss between the outputs of the decoder and the original Transformer input id representation of the entries (see Equations \ref{eq:decoderloss} and \ref{eq:decoderloss2}).
\begin{equation}
\label{eq:decoderloss}
  \begin{aligned}
      L(x, y) = \sum_{n=1}^{N}l_n
  \end{aligned}
\end{equation}
\begin{equation}
\label{eq:decoderloss2}
  \begin{aligned}
      l_n & = -w_{y_n}x_{n,y_{n}}, \\
          w_c = weight[c]&\cdot\mathbbm{1} \{c\neq ignore\_index\}
  \end{aligned}
\end{equation}

\subsection{Experimental Setup}
The optimizer used for our models is represented by AdamW~\cite{kingma2017adam}. The learning rate is set to \textit{2e-5}, while the loss functions used for the complexity task are the L1 loss~\cite{janocha2016loss} for the CompLex LCP dataset and the Mean Squared Error (MSE) loss~\cite{klined2005} for the CWI dataset. The auxiliary losses are summed to the main loss (i.e., complexity prediction) and are scaled according to their priority, with a factor of $ \alpha $, where $ \alpha $ is set to 0.1 for the VAE loss, and 0.01 for the Transformer decoder and task discriminator losses. The $\lambda$ parameter used for domain adaptation was updated according to Equation ~\ref{eq:lambdaeq}:
\begin{equation}
\label{eq:lambdaeq}
  \begin{aligned}
     \lambda=\frac{2}{1+e^{-\gamma \epsilon}} - 1
  \end{aligned}
\end{equation}
where $\epsilon$ is the number of epochs the model was trained; $\gamma$ was set to 0.1, while $ \beta $ was set to 0.2. Moreover, each model was trained for 8 epochs, except for the one including the VAE features, which was trained for 12 epochs.

\section{Results}
\subsection{LCP 2021 CompLex Dataset}

\begin{table*}[!ht]
\small
\centering
\caption{Results on the LCP 2021 English dataset.}
\begin{threeparttable}
\begin{tabular}{|l|c|c|c|c|c|c|c|c|}

\hline 
\multirow{3}{*}{\textbf{Model}} & \multicolumn{4}{c|}{\bf{Single-Word Target}} & \multicolumn{4}{c|}{\bf{Multi-Word Target}} \\ \cline{2-9}
& \multicolumn{2}{c|}{\bf{Trial}} & \multicolumn{2}{c|}{\bf{Test}} & \multicolumn{2}{c|}{\bf{Trial}} & \multicolumn{2}{c|}{\bf{Test}} \\ \cline{2-9}
& \bf Pearson & \bf MAE & \bf Pearson & \bf MAE & \bf Pearson & \bf MAE & \bf Pearson & \bf MAE  \\ \hline 
\newcite{dec3sl} & - & - & .4598 & .0866 & - & - & .3941  &  .1145 \\ \hline
\newcite{zaharia2021upb} & .7702 & .0671  & .7324 & .0677 & .7227 & .0863 & .7962   &  .0754 \\ \hline
\makecell[l]{1\textsuperscript{st} Place, \\LCP 2021~\cite{shardlow2021semeval}} & - &  -  & \textbf{.7886}  & \textbf{.0609} & - & - &  \textbf{.8612}  &  \textbf{.0616}  \\ \hline
\hline
\makecell[l]{Base (RoBERTa + Char BiLSTM)} & .7987 & .0654 & .7502 & .0682 & .7565 & .0828 & .8138 & .0739 \\ \hline
\makecell[l]{Base + DA} & .8111 & .0660 & .7569 & .0657 & \textbf{.7900} & \textbf{.0724} & .8246 & .0699 \\ \hline
\makecell[l]{Base + VAE + DA} & .8010 & .0658 & .7554 & .0669 & .7919 & .0745 & .8167 & .0761 \\ \hline
\makecell[l]{Base + Decoder + DA} & .7969 & .0687 & .7542 & .0704 & .7747 & .0812 & .8252 & .0693 \\ \hline
\makecell[l]{Base + Text simplification + DA} & \textbf{.8170} & \textbf{.0648} & .7744 & .0652 & .7670 & .0787 & .8285 & .0708 \\ \hline
\end{tabular}
\begin{tablenotes}\footnotesize
\item[*] DA = Domain Adaptation; VAE = Variational AutoEncoder; Decoder = Transformer Decoder; Pearson = Pearson Correlation Coefficient; MAE = Mean Absolute Error.\\
\end{tablenotes}
\end{threeparttable}
\label{tab:table1}
\end{table*}

\begin{table*}[!ht]
\small
\centering
\caption{Results on the CWI 2018 multilingual validation dataset.}
\begin{threeparttable}
\begin{tabular}{|p{3cm}|C{0.8cm}|C{0.8cm}|C{0.8cm}|C{0.8cm}|C{0.8cm}|C{0.8cm}|C{0.8cm}|C{0.8cm}|C{0.8cm}|C{0.8cm}|}

\hline 
\multirow{2}{*}{\textbf{Model}} & \multicolumn{2}{c|}{\bf{EN-N}} & \multicolumn{2}{c|}{\bf{EN-WN}} & \multicolumn{2}{c|}{\bf{EN-W}} & \multicolumn{2}{c|}{\bf{DE}} & \multicolumn{2}{c|}{\bf{ES}} \\ \cline{2-11}
& \bf P & \bf MAE & \bf P & \bf MAE & \bf P & \bf MAE & \bf P & \bf MAE & \bf P & \bf MAE  \\ \hline

\makecell[l]{Base (XLM-RoBERTa \\ + Char BiLSTM)} & .8517 & .0476 & .8460 & \textbf{.0512} & .7640 & \textbf{.0697} & .7092 & .0559 & \textbf{.6944} & .0635 \\ \hline
\makecell[l]{Base + LA} & .8592 & .0468 & .8431 & .0532  & .7773 & .0702 & .6857 & \textbf{.0551} & .6868 & \textbf{.0625} \\ \hline
\makecell[l]{Base +  VAE + LA} & .8557 & \textbf{.0463} & .8376 & .0527 & .7562 & .0702 & .7026 & .0565 & .6805 & .0628 \\ \hline
\makecell[l]{Base +  Decoder +  LA} & .8492 & .0511 & .8273 & .0569 & .7619 & .0745 & .6823 & .0645 & .6519 & .0725 \\ \hline
\makecell[l]{Base +  Text \\ simplification +  LA} & \textbf{.8602} & .0514 & \textbf{.8555} & .0560 & \textbf{.7842} & .0716 & \textbf{.7147} & .0621 & .6787 & .0688 \\ \hline
\end{tabular}
\begin{tablenotes}\footnotesize
\item[*] LA = Language Adaptation; VAE = Variational AutoEncoder; Decoder = Transformer Decoder; EN-N = English-News; EN-WN = English-WikiNews; EN-W = English-Wikipedia; DE = German; ES = Spanish; P = Pearson Correlation Coefficient; MAE = Mean Absolute Error.\\
\end{tablenotes}
\end{threeparttable}
\label{tab:table2}
\end{table*}

\begin{table*}[!ht]
\small
\centering
\caption{Results on the CWI 2018 multilingual test dataset.}
\begin{threeparttable}
\begin{tabular}{|p{2.8cm}|C{0.62cm}|C{0.62cm}|C{0.62cm}|C{0.62cm}|C{0.62cm}|C{0.62cm}|C{0.62cm}|C{0.62cm}|C{0.62cm}|C{0.62cm}|C{0.62cm}|C{0.62cm}|}

\hline 
\multirow{2}{*}{\textbf{Model}} & \multicolumn{2}{c|}{\bf{EN-N}} & \multicolumn{2}{c|}{\bf{EN-WN}} & \multicolumn{2}{c|}{\bf{EN-W}} & \multicolumn{2}{c|}{\bf{DE}} & \multicolumn{2}{c|}{\bf{ES}}  & \multicolumn{2}{c|}{\bf{FR}} \\ \cline{2-13}
& \bf \scriptsize P & \bf \scriptsize MAE & \bf \scriptsize P & \bf \scriptsize MAE & \bf \scriptsize P & \bf \scriptsize MAE & \bf \scriptsize P & \bf \scriptsize MAE & \bf \scriptsize P & \bf \scriptsize MAE & \bf \scriptsize P & \bf \scriptsize MAE \\ \hline
\newcite{kajiwara-komachi-2018-complex} & - & .0510 & - & .0704 & - & .0931 & - & .0610 & - & .0718 & - & .0778  \\ \hline
\newcite{BingelBjerva2018} & - & - & - & - & - & -  & - & .0747 & - & .0789 & - & \textbf{.0660}   \\ \hline
\newcite{gooding-kochmar-2018-camb}  & - & .0558 & - & .0674 & - & .0739   & - & - & - & - & - & -   \\ \hline
\hline
\makecell[l]{Base \\(XLM-RoBERTa \\ + Char BiLSTM)} & .8560 & .0461 & .8045 & .0533 & .7205 & .0679 & \textbf{.7405} & \textbf{.0540} & .6873 & .0619 & .5506 & .0793 \\ \hline
\makecell[l]{Base + LA} & \textbf{.8582} & .0466 & .8146 & \textbf{.0513} & .7310 & .0700 & .6866 & .0558 & .6809 & .0606 & .5409 & .0842 \\ \hline
\makecell[l]{Base +  VAE + LA} & .8580 & \textbf{.0450} & .8060 & .0526 & .7354 & \textbf{.0671} & .7131 & .0553 & \textbf{.6912} & \textbf{.0595} & \textbf{.5559} & .0752 \\ \hline
\makecell[l]{Base + Decoder +  LA} & .8533 & .0509 & .7978 & .0560 & .7124 & .0708 & .6976 & .0653 & .6490 & .0692 & .4663 & .0889 \\ \hline
\makecell[l]{Base +  Text \\ simplification +  LA} & .8580 & .0502 & \textbf{.8338} & .0539 & \textbf{.7420} & .0707 & .7230 & .0614 & .6837 & .0671 & .5394 & .0876\\ \hline
\end{tabular}
\begin{tablenotes}\footnotesize
\item[*] LA = Language Adaptation; VAE = Variational AutoEncoder; Decoder = Transformer Decoder; EN-N = English-News; EN-WN = English-WikiNews; EN-W = English-Wikipedia; DE = German; ES = Spanish; FR = French; P = Pearson Correlation Coefficient; MAE = Mean Absolute Error.\\
\end{tablenotes}
\end{threeparttable}
\label{tab:table3}
\end{table*}

We consider as baselines two models used for the LCP 2021 competition~\cite{shardlow2021semeval}, as well as the best-registered score. \newcite{dec3sl} employed the usage of neural network solutions; more specifically, they used chunks of the sentences obtained with Sent2Vec as input features. \newcite{zaharia2021upb} created models that are based on target and context feature extractors, alongside features resulted from Graph Convolutional Networks, Capsule Networks, and pre-trained word embeddings.

Table \ref{tab:table1} depicts the results obtained for the English dataset using domain adaptation and various configurations. "Base" denotes the initial model (RoBERTa + Char BiLSTM) on which we apply domain adaptation, as well as the auxiliary tasks. The domain adaptation technique offers improved performance when applied on top of an architecture, considering that the model learns cross-domain features. The only exception is represented by a slightly lower Pearson score on the model that uses domain adaptation alongside the Transformer decoding auxiliary task (Base + Decoder + DA), with a value of .7969 on the trial dataset, when compared to the initial .7987 (Base). However, the remaining models improve upon the starting architecture, with the largest improvements being observed for domain adaptation and the text simplification auxiliary task (Base + Text simplification + DA), with a Pearson correlation coefficient on the test dataset of .7744, 2.42\% better than the base model. The improved performance can be also seen for the Mean Absolute Error score (MAE = .0652).

While the Transformer decoder auxiliary task does not offer the best performance for the single word dataset, the same architecture offers the second-best performance for the multi-word dataset, with a Pearson score of .8252 compared to the best one, .8285. The domain adaptation and VAE configuration provide improvements upon the base model (.7554 versus .7502 Pearson), but the VAE does not have an important contribution, considering that the Base + domain adaptation model has a slightly higher Pearson score of .7569. 

\subsection{CWI 2018 Dataset}

We also experimented with a multilingual dataset, where the discriminant is considered to be the language. The baseline consists of three models used from the CWI 2018 competition. The performance is evaluated in terms of MAE; however, we also report the Pearson Correlation Coefficient. First, \newcite{kajiwara-komachi-2018-complex}  based their models on regressors, alongside features represented by the number of characters or words and the frequency of the target word in certain corpora. Second, the approach of \newcite{BingelBjerva2018} is based on Random Forest Regressors, as well as feed-forward neural networks alongside specific features, such as log-probability, inflectional complexity, or target-sentence similarity; the authors focused on non-English entries. Third, \newcite{gooding-kochmar-2018-camb} approach the English section of the dataset by employing linear regressions. The authors used several types of handcrafted features, including word n-grams, POS tags, dependency parse relations, and psycholinguistic features.

Table \ref{tab:table2} presents the results obtained on the multilingual validation dataset and compares the performance of different configurations. The best overall performance in terms of Pearson correlation coefficient is yielded by the Base model (XLM-RoBERTa + Char BiLSTM) alongside the text simplification auxiliary task and the domain adaptation technique (Base + Text simplification + LA), with values of .8602 on English News, .8555 on English WikiNews, as well as .7842 on English Wikipedia and .7147 on German. The best Pearson score for the Spanish language is obtained by the base model, with .6944. The Base + VAE + LA architecture offers improvements over the Base model, but falls behind when compared to the Base + Text simplification + LA model, with Pearson correlation ranging from .8557 on the English News dataset to .6805 on the Spanish dataset.

However, when switching to MAE, the metric used for evaluation in the CWI 2018 competition, the best performance is split between the first three models, namely Base, Base + LA, and Base + VAE + LA. The Base + LA approach yields the best, lowest MAE score on the German and Spanish datasets, while the Base architecture performs the best on English WikiNews and English Wikipedia. The English News achieves the best MAE results from the Base + VAE + LA model.

Nevertheless, the best overall performance is obtained by the Base + VAE + LA model on the test dataset (see Table \ref{tab:table3}), with dominating Pearson and MAE scores on the Spanish and French languages: 0.6912 Pearson, 0.595 MAE for Spanish, as well as .5559 Pearson, and .0752 MAE for French, respectively. The Base + Text simplification + LA model performs the best in terms of Pearson Correlation Coefficient on the English WikiNews and Wikipedia datasets, with Pearson scores of .8338 and .7420. However, the best MAE scores for the same datasets are generated by the Base + LA model (.0513 English WikiNews) and Base + VAE + LA (.0671 English Wikipedia).

\section{Discussions}
The domain adaptation technique supports our model to learn general cross-domain or cross-language features, while achieving higher performance. Moreover, jointly training on two different tasks (i.e., lexical complexity prediction and text simplification), coupled with domain adaptation to generalize the features from the two tasks, can lead to improved results. 

However, there are entries for which our models were unable to properly predict the complexity score, namely: a) entries with a different level of complexity (i.e. biomedical), and b) entries part of a language that was not present in the training dataset (i.e., French). For the former, scientific terms (e.g., "sitosterolemia"), abbreviations (e.g., "ES"), or complex elements (e.g., "H3-2meK9") impose a series of difficulties for our feature extractors, considering the absence of these tokens from the Transformer vocabulary. The latter category of problematic entries creates new challenges in the sense that it represents a completely new language on which the architecture is tested. However, as seen in the results section, the cross-lingual domain adaptation technique offers good improvements, helping the model achieve better performance on French, even though the initial architecture was not exposed to any French example.

\section{Conclusions and Future Work}
This work proposes a series of training techniques, including domain adaptation, as well as multi-task adversarial learning, that can be used for improving the overall performance of the models for CWI. Domain adaptation improves results by encouraging the models to extract more general features, that can be further used for the lexical complexity prediction task. Moreover, by jointly training the model on the CWI tasks and an auxiliary similar task (i.e., text simplification), the overall performance is improved. The task discriminator also ensures the extraction of general features, thus making the model more robust on the CWI dataset.

For future work, we intend to experiment with meta-learning~\cite{finn2017modelagnostic} alongside domain adaptation ~\cite{WangKe8878083}, considering the scope of the previously applied training techniques. This would enable us to initialize the model's weights in the best manner, thus ensuring optimal results during the training phase.

\section*{Acknowledgments}
This research was supported by a grant of the Romanian National Authority for Scientific Research and Innovation, CNCS - UEFISCDI, project number TE 70 PN-III-P1-1.1-TE-2019-2209, "ATES - Automated Text Evaluation and Simplification".

\bibliographystyle{acl_natbib}
\bibliography{anthology,bibliography}

\end{document}